# Improving Fine-Tuning with Latent Cluster Correction

Cédric Ho Thanh
*Advanced Data Science Project, RIKEN Information R&D and Strategy Headquarters*
*RIKEN*
Tokyo, Japan
cedric.hothanh@riken.jp

*Abstract*—The existence of salient semantic clusters in the latent spaces of a neural network during training strongly correlates its final accuracy on classification tasks. This paper proposes a novel fine-tuning method that boosts performance by optimising the formation of these latent clusters, using the Louvain community detection algorithm and a specifically designed clustering loss function. We present preliminary results that demonstrate the viability of this process on classical neural network architectures during fine-tuning on the CIFAR-100 dataset.

## I. Introduction

### A. Background

Neural networks (NNs) are a family of machine learning models obtained by composition of simple, usually learnable operations called *layers*. In its simplest form, a NN $f$ can be expressed as

$$f(x) = f_n(f_{n-1}(\cdots f_1(x) \cdots)), \qquad (1)$$

where $x$ is an input sample, and where $f_1, ..., f_n$ are $f$'s layers, e.g. fully connected, convolutional, dropout, softmax, ReLU… Depending on the application, $x$ can represent an abstract vector of numbers, an image, an electrocardiogram, etc. The present paper deals with *classifier NNs* which places an input $x$ in a discrete category by assigning a predicted label $y_{\text{pred}} = f(x)$ to it.

A *latent representation* (LR) of $x$ is any intermediate value $z = f_k(\cdots f_1(x) \cdots)$ for some $1 \leq k < n$, obtained while evaluating the NN on $x$. The space in which $z$ lives, i.e. the output space of $f_k$, is called a *latent space* (LS). Although the relationship between $x$ and its predicted label $y_{\text{pred}} = f(x)$ can be somewhat understood, the relationship between $x$, $y$, and the LRs of $x$ is more mysterious.

During training, a NN learns to extract features from the input $x$ and represents them in its abstract LSs, shuffles them in a way that is (hopefully) beneficial by passing the representations through further transformation layers, and finally make a classification decision. However the precise way in which the NN decides to allocate its LSs is not usually known, and perhaps not even *knowable*. Indeed, modern NNs can have dozens of different LSs each spanning hundreads of thousands of dimensions!

Nonetheness, it is understood that eventhough the dimension of a chosen LS can be large, the actual (or *intrinsic*) number dimensions used by the NN can be much smaller, and some loose patterns may emerge. Consider Fig. 1 representing the structures of several LSs[1] of an instance of ResNet-18 [1] trained on the CIFAR-10 dataset [2].

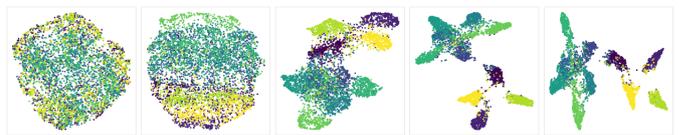

Fig. 1: LRs of 5000 samples from the CIFAR-10 dataset passed through a pretrained instance of ResNet-18, after dimensionality reduction using UMAP [3], coloured by class. The LS sits deeper in the network as we go from left (feature extraction layers) to right (classification layers).

It is apparent that as we go deeper into the network (i.e. apply more and more layers to the input), the successive LSs become increasingly structured. Representations of samples from the same class, while initially scattered during the feature extraction phase, get increasingly clumped, or *clustered* together, while the clusters get increasingly *separated*.

With this in mind, a few natural questions emerge:
1) how does LS structure (mainly clustering and separation) relate to the NN's performance?
2) how to measure the quality of LS structures?
3) how to influence LS structure during training to improve the NN's performance?

### B. Related works

Separation and clustering in LSs has been investigated in several works.

First, [4] introduces the *generalized discrimination distance* (GDV) which is perhaps the most direct way to measure separation and clustering of a set of samples. Through empirical study, they demonstrate that GDV is relevant for the study of LSs of multi-layer perceptrons (MLPs).

In [5], the authors place more emphasis in the study of the geometry (such as the dimension and radius) of the *object manifolds*, which are the manifolds defined by the LR of samples in a given class. They investigate the capacity of a NN to linearly separate these manifolds by computing a quantity they call *system load*. Their empirical study shows that learning tends to increase separability and decrease the dimension and radius of object manifolds.

---

[1] We shall use the expression "LS" a little loosely to also refer to the *distribution* of LRs in the LS. So by "structure of the LS" we may also refer to the patterns that emerge in the distribution of LRs.

In [6] and [7], the authors try to actively influence the layout of the LSs by introducing the *graph smoothness loss*. That graph is constructed from the LRs, where the vertices are the representations and the edges are weighted inversely exponentially by distance. By training a NN to exclusively minimize this graph smoothness loss, and then training a separate small classifier model (either a MLP or a support vector machine) on top of that, the authors achieve performances similar to NNs trained classically using cross-entropy loss. In [7] specifically, the authors also use these separations methods for knowledge distillation ([8], [9]) and to improve the robustness of classifier models against dataset corruption or adversarial attacks.

The works of [10] go even further by designing NN architectures that are more amenable to latent separation. In their first approach, the authors create a model consisting of $M$ parallel deep convolutional neural networks (DCNN), where $M$ is the number of classes. Each is a "one-vs-rest" binary classifier for the corresponding class tasked with clustering LRs of its class away from samples from other classes. The second approach is more involved. First, a single DCNN is trained to maximize the *Grassmanian geodesic distance* (GGD) [11] between the subspaces of LR of each classes. Unlike traditional separation metrics that aim to minimize the distance between samples of the same class and/or maximize the distance between samples of different classes, the GGD merely encourages the network to allocate different linear latent subspaces for each class in such a way that the *principal angles* between these subspaces are as large as possible.

The aforementionned works, along with the present paper, only deal with classifier NNs. In [12], the author applies these separation and clustering ideas to autoencoders (AEs). Instead of naively encouraging clusters in the LS² of an AE, an extra NN, called *representation network* is attached to the encoder alongside the decoder. The role of the representation network is to embed LRs to another space in which a clustering metric is applied.

*C. Contribution*

The present work follows the same idea as many of the papers above, namely encouraging clustering to improve performance. Our method, which we call *latent cluster correction* (LCC), uses a nearest-neighbor-based Louvain community detection algorithm to find these clusters (Section III.A), an assignment method to interpret their "intended" class (Section III.C), and a new loss function to encourage the formation of more salient and accurate clusters (Section III.D).

*D. Acknowledgments*

We would like to thank the anonymous reviewers of the FITML'24 workshop for their constructive feedback. We would also like to thank Prof. Jun Seita and Dr. Dorothy Ellis for their helpful guidance and discussions. This research was supported by RIKEN Pioneering Project "Prediction for Science".

## II. Choice of clustering algorithm

The method by which latent clusters are detected is crucial to the success of the LCC.

Although the final latent structure of Fig. 1 seems simple enough to be solvable via an ensemble support vector machine scheme, one must keep in mind that the process of dimensionality reduction inevitably destroys a lot of information. A cluster in the 2D representation may in fact contain several clusters in the original high-dimensional LS, and the relative position and shape of 2D clusters does not reflect the true latent geometry in general. Furthermore, empirical evidence suggests that a LS may exhibit more clusters than there are true classes.

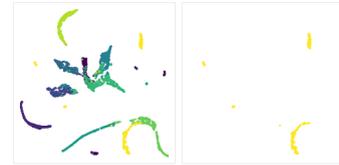

Fig. 2: On the left: LRs of 5000 samples from the Fashion MNIST [13] dataset passed through a pretrained instance of AlexNet [14], after dimensionality reduction using UMAP, coloured by class. On the right: same, but only the LRs belonging to the "Ankle boot" class are plotted.

For example, the right plot of Fig. 2 displays the (dimensionality-reduced) LRs of a single class. Two significant clusters are visible. This separation originates from the presence of multiple (non-dimensionality-reduced) clusters among LRs of this class. We can think of this phenomenon as the NN deciding to segregate samples in the same class based on some extracted features, only to ultimately assign the same label to them. Therefore, we argue that forcing the NN to make only one cluster per class, as is the case in [6] and [10] among others, could be counterproductive.

In particular, the algorithm used for detecting the clusters should not have the expected number of clusters as a hyperparameter, which rules out $k$-means for example. In fact, the choice of algorithm and its hyperparameters should carry as few empirical biases as possible due to the lack of understanding of the latent geometry. For example, the clustering algorithm should be able to deal with non-linearly separable clusters, shouldn't produce outliers, should produce meaningful clusters even if the data is high-dimensional, etc. Table I compares a small selection of the most common clustering algorithms against our requirements. This list is far from exhaustive, and a more complete account can be found in [15], [16], and [17].

---

²When talking about AEs, the term "LS" refers only to the space that sits between the encoder and decoder.

| Algorithm | Handles unknown nb. of clsts. | Handles non-linearly sep. clsts. | Few params. | No outliers |
|---|---|---|---|---|
| $k$-means | × | × | ✓ | ✓ |
| Gaussian mixture | × | × | ✓ | ✓ |
| BIRCH [18] | ✓ | × | ✓ | ✓ |
| CLIQUE [19] | ✓ | ✓ | ✓ | ✓ |
| OPTICS [20] | ✓ | ✓ | × | × |
| Spectral clustering [21] | × | ✓ | × | ✓ |
| Affinity propagation [22] | ✓ | × | ✓ | ✓ |
| HDBSCAN [23] | ✓ | ✓ | × | × |
| $k$-NN-Louvain (used in this work) | ✓ | ✓ | ✓ | ✓ |
| $k$-NN pressure (See Section VI.A) | ✓ | ✓ | ✓ | ✓ |

Our method, $k$-NN-Louvain, first constructs the $k$-nearest neighbors ($k$-NN) graph of the dataset of LRs, and then applies the Louvain method to find *communities* (synonymous with "cluster") in the graph, which correspond to clusters in LS. Other flexible graph community detection algorithm exist and could be used in principle, see Section VI.A, and [24] for a survey. Using these alternative methods is the subject of ongoing studies.

## III. THE CLUSTERING LOSS

The *clustering loss* measures the quality and accuracy (*vis-à-vis* the true labels) of the clusters in a chosen LS of a NN $f$. If $f(x) = f_n(f_{n-1}(\cdots f_1(x)\cdots))$ and we selected an index $1 \leq k < n$, then the clustering loss of $f$ on a dataset $(X, y)$ is computed in four steps:

1) compute the dataset of LRs $Z = f_k(\cdots f_1(X) \cdots)$;
2) compute the latent clusters of $Z$ using our proposed $k$-NN-Louvain method:
   a) (Section III.A) compute the $k$-NN graph $G$ of $Z$;
   b) (Section III.B) find the Louvain communities of $G$ which creates a new vector $y_{\text{clst}}$ of *clusters*;
3) (Section III.C) find an optimal *one-to-many matching* between the true labels in $y$ and $y_{\text{clst}}$;
4) (Section III.D) based on this matching, find the *misclustered* latent samples and compute the loss term associated to each; the clustering loss $\mathcal{L}_{\text{clst}}$ is the mean of all these terms.

### A. The Louvain-Leiden community detection algorithm

Given a weighted undirected graph $G = (V, E)$, what is the best partition of $V$ into *communities* so as to maximise the weighted connectivity within each community while minimizing it in between communities? A popular approximation method known as the *Louvain algorithm* [25] seeks to maximize a surrogate objective called *modularity*, given by

$$Q = \frac{1}{2m} \sum_{i,j} \left( A_{i,j} - \frac{k_i k_j}{2m} \right) \delta_{i,j} \qquad (2)$$

where $A$ is the adjacency matrix of $G$, $m$ is the sum of all edge weights, $k_i$ is the sum of the weights of all edges incident to node $i$, and $\delta_{i,j}$ is 1 if node $i$ and $j$ are in the same community, and 0 otherwise. The algorithm then produces a series of partitioned graphs $G = G^{(1)}, G^{(2)}, \ldots$ as follows.

1) (initialization) Start with the trivial partition where every node belongs to its own community.
2) (modularity optimization) For every edge $(i, j) \in E$, we consider whether is it beneficial to remove node $i$ from its current community and move it to the community of its neighbor $j$, say $C \subseteq V$. The benefit is measured by the *modularity gain* $\Delta Q$, which is a quantity that depends on the connectivity within $C$, and between $C$ and $i$. Node $i$ is moved to the community of its neighbor $j$ for which the modularity gain is the largest, provided it is positive. If all modularity gains are negative, node $i$ does not move.
3) (aggregation) A new graph $G^{(i+1)}$ is built, where
   a) the nodes are the communities of $G^{(i)}$ produced by step 2,
   b) the adjacency matrix of $G^{(i+1)}$ is given by $A_{C,D}^{(i+1)} = \sum_{c \in C, d \in D} A_{c,d}^{(i)}$ (i.e. the total weight of edges between $C$ and $D$) if $C \neq D$, and $A_{C,C}^{(i+1)} = \frac{1}{2} \sum_{c,d \in C} A_{c,d}^{(i)}$ (i.e. the total weight of edges in $C$).

These steps are repeated until no further improvement is possible, i.e. until a graph $G^{(n)}$ is produced for which no "node moving" in step 2 offers a positive modularity gain. At this point, nodes of $G^{(n)}$ can be traced back to a partition of the nodes of $G$ into disjoint communities. Unfortunately, the Louvain algorithm sometimes produces sparse communities. The *Louvain-Leiden algorithm* [26] is an improvement that prevents this. It works by adding a step between step 1 and 2 that further partitions communities into subcommunities following a set of complex rules. Then, at the beginning of the next iteration, instead of starting out with the trivial partition (that in which every node belongs to its own community), subcommunities of the same community are already grouped together. In the sequel, we refer to the Louvain-Leiden algorithm as simply "Louvain algorithm".

### B. Latent clusters

Let $(Z, y)$ be a labeled dataset of LRs, where $Z = (z_1, \ldots, z_N)^t \in \mathbb{R}^{N \times d}$. The integer $d$ is called the *latent dimension*.

The first step towards computing the clustering loss of $(Z, y)$ is to construct the $k$-nearest neighbor ($k$-NN) graph $G$ of $Z$, where $k$ is a hyperparameter chosen in advance. The set of nodes of $G$ is simply $\{z_i \mid 1 \leq i \leq N\}$, and nodes $z_i$ and $z_j$ are linked if $z_i$ is among the $k$-NNs of $z_j$ or conversely.[3] In this case, the weight of the edge is given by the Gaussian radial basis function $\exp(-\|z_i - z_j\|^2)$.

---
[3]An element is not considered to be one of its own neighbors.

Then, running the Louvain algorithm on $G$ produces a partition of $Z$ into communities $C_1, C_2, ...$, which gives rise to the *vector of clusters* $y_{\text{clst}} \in \mathbb{N}^N$, where $y_{\text{clst},i} = j$ if $z_i \in C_j$. It is expected that most of the time $y \neq y_{\text{clst}}$ even up to label permutation, and in fact, the two vectors might not even have the same number of distinct labels (see Fig. 2).

### C. True labels vs. clusters

To establish a relationship between $y$ and $y_{\text{clst}}$, we construct an *assignment* between true labels and clusters. To disambiguate, let us denote the set of true labels (i.e. the distinct values of $y$) by $L_{\text{true}} \subseteq \mathbb{N}$, and the set of clusters (indices) by $L_{\text{clst}} \subseteq \mathbb{N}$. By changing the names as necessary, we may assume that $L_{\text{true}} \cap L_{\text{clst}} = \emptyset$.

A *one-to-many matching* between $L_{\text{true}}$ and $L_{\text{clst}}$ is simply a function $\alpha : L_{\text{clst}} \to L_{\text{true}}$. We think of the cluster $c \in L_{\text{clst}}$ as "being matched" to the true label $\alpha(c) \in L_{\text{true}}$. The assignment is optimal if the following value is maximized:

$$\sum_{t \in L_{\text{true}}} \left| \{z_i \mid y_i = t\} \cap \{z_i \mid \alpha(y_{\text{clst},i}) = t\} \right|. \quad (3)$$

In other words, clusters are assigned to true labels so as to minimize the overall discrepency between the set of samples with true labels $t \in L_{\text{true}}$ and the set of samples in clusters matched to $t$.

Such an optimal assignment can be found by reformulating the objective function of (3) as a *discrete max-flow-max-weight problem*. A directed graph $F = (V', E')$ is constructed, where the underlying set of nodes is

$$V' = \{\top, \bot\} \cup L_{\text{true}} \cup L_{\text{clst}}. \quad (4)$$

Node $\top$ is called the *supersource* whereas $\bot$ is the *supersink*. The edges and their weight and *capacity* are given by the following rules:
1) for all $t \in L_{\text{true}}$, there is an edge from $\top$ to $t$ whose weight is 0 and whose capacity is infinite;
2) for all $t \in L_{\text{true}}$ and all $c \in L_{\text{clst}}$, there is an edge from $t$ to $c$ whose weight is

$$\left| \{z_i \mid y_n = t\} \cap \{z_i \mid y_{\text{clst},i} = c\} \right| \quad (5)$$

and whose capacity is 1;
3) for all $c \in L_{\text{clst}}$, there is an edge from $c$ to $\bot$ whose weight is 0 and whose capacity is 1.

A *discrete flow* from $\top$ to $\bot$ is a function $g : E' \to \mathbb{N}$ such that:
1) for all node $v$ except $\top$ and $\bot$,

$$\sum_{(w,v) \in E} g(w, v) = \sum_{(v,w) \in E'} g(v, w), \quad (6)$$

i.e. the incoming flow of node $v$ equals its outgoing flow;
2) for all edge $e \in E'$, $g(e)$ is at most the capacity of $e$.

The flow is optimal if its *weight* $\sum_{e \in E'} g(e)w(e)$ is maximized, where $w(e)$ is the weight of edge $e$.

**Proposition** Let $g$ be an optimal flow. For a cluster $c \in L_{\text{clst}}$, there exists a unique true label $t_c \in L_{\text{true}}$ such that $g(t_c, c) > 0$. Furthermore, $g$ gives rise to an optimal assignment $\alpha : L_{\text{clst}} \to L_{\text{true}}$ where every $c \in L_{\text{clst}}$ is mapped to its corresponding $t_c \in L_{\text{true}}$ defined above.

**Proof** Assume that such a $t_c$ does not exist. In particular the outgoing flow of $c$ is 0. For any choice of $t \in L_{\text{true}}$, it is thus possible to add a flow of 1 from $t$ to $c$ and from $c$ to $\bot$ without violating any flow constraint. Since all edge weights are nonnegative, the flow weight would increase, contradicting the optimality of $g$. Next, assume that there is another $t' \neq t$ such that $g(t', c) > 0$. This means that the incoming flow of $c$ is greater than 1, and therefore the outgoing flow $g(c, \bot)$ is also greater than 1. This is not possible since the capacity of the edge $(c, \bot)$ is 1. Finally, optimality of $\alpha$ in the sense of (3) follows from the optimality of $g$. $\square$

### D. The clustering loss

At this stage, we dispose of a cluster vector $y_{\text{clst}}$ which regroups samples following an optimal $k$-NN clustering scheme. We also constructed a matching $\alpha : L_{\text{clst}} \to L_{\text{true}}$ which assigns a true label to each cluster.

We say that a sample $z_i$ is *misclustered* (MC) if $\alpha(y_{\text{clst},i}) \neq y_i$, i.e. its cluster $y_{\text{clst},i}$ is not assigned to its true label $y_i$. It is *correctly clustered* (CC) otherwise. We say that a MC sample $z_i$ is *correctible* if there exists $k$ CC samples $z_{j_1}, ..., z_{j_k}$ in $y_i$. In this case, the *target* of $z_i$ is the centroid

$$\bar{z}_i = \frac{z_{j_1} + \cdots + z_{j_k}}{k}, \quad (7)$$

where $z_{j_1}, ..., z_{j_k}$ are the $k$ nearest neighbors of $z_i$ among the CC samples in $y_i$.

The *clustering loss* of the dataset of LRs $(Z, y)$ is defined as

$$\mathcal{L}_{\text{clst}} = \frac{1}{\sqrt{d} N_{\text{corr}}} \sum_{z_i \text{ corr.}} \|z_i - \bar{z}_i\| \quad (8)$$

where $d$ is the latent dimension and $N_{\text{corr}}$ is the number of correctible samples.

## IV. EXPERIMENTS

### A. Setup

This preliminary study focuses on three model architectures, varying from small to somewhat large.

| Model | Nb. of weights |
|---|---|
| TinyNet [27], [28] | $\approx 2.06 \times 10^6$ |
| ResNet–18 [1] | $\approx 11.2 \times 10^6$ |
| AlexNet [14] | $\approx 57.4 \times 10^6$ |

All three were pretrained on the ImageNet dataset [29], and the task at hand is to fine-tune them on the CIFAR-100 dataset [2] using LCC.

The selected layers for correction were either the last fully-connected layer (which outputs the logits, also called the classifier *head*), or the second to last trainable layer. The clustering loss weight $w$ ranged over $\{10^{-2}, 10^{-4}\}$. The number $k$ of neighbors considered for clustering ranged over

{5, 50, 500}. Each trial started with a single *warmup* epoch where the clustering loss was not applied.

Fine-tuning was performed using stochastic gradient descent with momentum $\mu = 0.9$ and a cosine learning rate scheduling over 50 epochs, starting at $\gamma = 0.1$.

### B. Results

The best parameters and results are summarized in the tables below. The first row of every table corresponds to the baseline, which has been fine-tuned without LCC.

Applying LCC on ResNet-18 (Table III) shows appreciable and consistent accuracy gains. The same is not true in AlexNet's case (Table IV), where the accuracy difference is overall negligible. This suggests that the impact of LCC is highly model-dependent.

TABLE III: Benchmark results for ResNet-18.

| Layer | $w$ | $k$ | Acc. | Gain |
|---|---|---|---|---|
| | | | 78.51% | |
| Head | $10^{-4}$ | 5 | 78.86% | +0.35% |
| 2$^{nd}$ to last | $10^{-4}$ | 5 | 78.99% | +0.48% |
| Head | $10^{-4}$ | 50 | 78.92% | +0.41% |
| 2$^{nd}$ to last | $10^{-4}$ | 50 | 79.05% | +0.54% |
| **Head** | $10^{-4}$ | 500 | **79.53%** | **+1.02%** |
| 2$^{nd}$ to last | $10^{-4}$ | 500 | 79.38% | +0.87% |
| Head | $10^{-2}$ | 5 | 79.24% | +0.73% |
| 2$^{nd}$ to last | $10^{-2}$ | 5 | 79.22% | +0.71% |
| Head | $10^{-2}$ | 50 | 79.1% | +0.59% |
| 2$^{nd}$ to last | $10^{-2}$ | 50 | 79.46% | +0.95% |
| Head | $10^{-2}$ | 500 | 79.03% | +0.52% |
| 2$^{nd}$ to last | $10^{-2}$ | 500 | 79.48% | +0.97% |

TABLE IV: Benchmark results for AlexNet.

| Layer | $w$ | $k$ | Acc. | Gain |
|---|---|---|---|---|
| | | | 71.49% | |
| Head | $10^{-4}$ | 5 | 71.31% | −0.18% |
| 2$^{nd}$ to last | $10^{-4}$ | 5 | 71.39% | −0.1% |
| Head | $10^{-4}$ | 50 | 71.29% | −0.2% |
| 2$^{nd}$ to last | $10^{-4}$ | 50 | 71.41% | −0.08% |
| Head | $10^{-4}$ | 500 | 71.19% | −0.3% |
| 2$^{nd}$ to last | $10^{-4}$ | 500 | 70.91% | −0.58% |
| Head | $10^{-2}$ | 5 | 71.69% | +0.2% |
| 2$^{nd}$ to last | $10^{-2}$ | 5 | 71.48% | −0.02% |
| **Head** | $10^{-2}$ | 50 | **71.72%** | **+0.23%** |
| 2$^{nd}$ to last | $10^{-2}$ | 50 | 71.46% | −0.03% |
| Head | $10^{-2}$ | 500 | 71.41% | −0.08% |
| 2$^{nd}$ to last | $10^{-2}$ | 500 | 71.1% | −0.39% |

### C. Accuracy of correctly clustered and misclustered samples

Let us consider the implication for a sample to be CC or MC *vis-à-vis* the classification accuracy. Consider a NN $f$ decomposed as $f(x) = f_n(f_{n-1}(\cdots f_1(x) \cdots))$. Let $(X, y)$ be a labeled dataset. For each $1 \leq k < n$ we consider the dataset of LRs $Z^k = f_k(\cdots f_1(X) \cdots)$.

Let $X^{k,\text{CC}}$ and $X^{k,\text{MC}}$ be the subsets of those input samples whose LR after $f_k$ is CC or MC respectively, and $y^{k,\text{CC}}$ and $y^{k,\text{MC}}$ be the corresponding label vectors. After any given pretraining and fine-tuning epoch of $f$, we observe that the following inequalities almost always hold:

$$\begin{aligned}\text{acc}(X^{k,\text{MC}}, y^{k,\text{MC}}) &< \text{acc}(X, y) < \text{acc}(X^{k,\text{CC}}, y^{k,\text{CC}}), \\ \text{ce}(X^{k,\text{MC}}, y^{k,\text{MC}}) &> \text{ce}(X, y) > \text{ce}(X^{k,\text{CC}}, y^{k,\text{CC}}),\end{aligned} \quad (9)$$

where acc stands for the accuracy of $f$ on a given labeled dataset, and ce for the cross-entropy loss. The discrepancy increases with $k$, i.e. as we go deeper in the NN.

Fig. 3 illustrate the accuracy inequalities. Here, $f$ is an instance of AlexNet being pretrained. Several LSs of $f$ have been selected and sorted by depth from left to right and top to bottom. Each curve represents the accuracy of $f$ at a given epoch, for a given subset of samples. The blue curves correspond to $\text{acc}(X^{k,\text{CC}}, y^{k,\text{CC}})$, red to $\text{acc}(X^{k,\text{MC}}, y^{k,\text{MC}})$, and dashed black to the global accuracy $\text{acc}(X, y)$.[4] See also Fig. 4 for a similar plot for accuracy during fine-tuning with LCC.

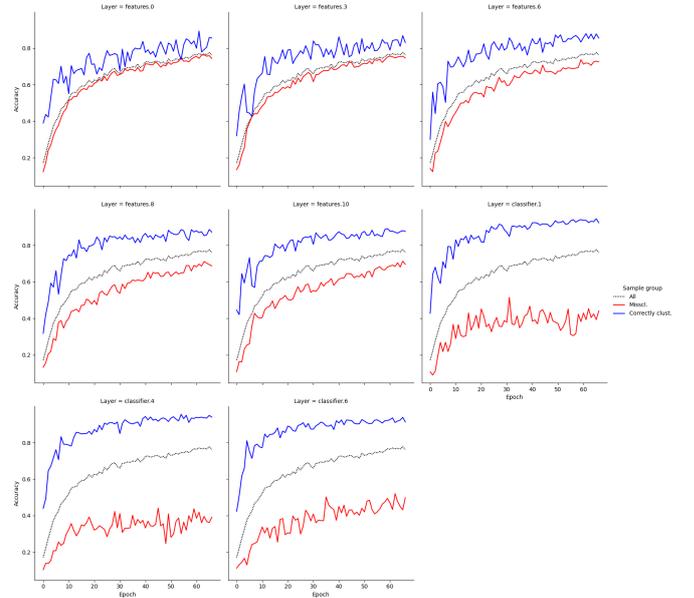

Fig. 3: Accuracy of correctly clustered and missclustered samples during the pretraining of AlexNet on CIFAR-10.

---
[4]The black curve is thus the same across all plots.

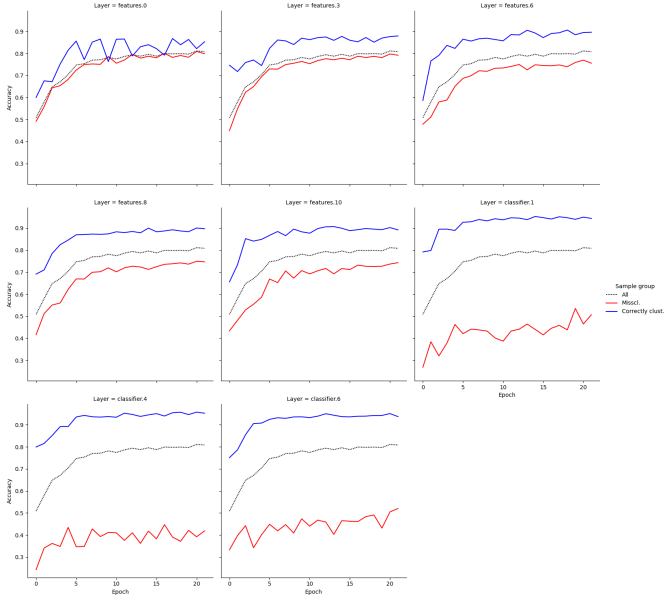

Fig. 4: Accuracy of correctly clustered and missclustered samples during the fine-tuning of AlexNet on CIFAR-10, with $k = 5$ and $w = 10^{-3}$.

## V. Discussion

### A. Computational cost

LCC comes at a steep computational cost. First, at the begining of each post-warmup fine-tuning epoch, the correction targets of MC samples have to be computed (see Section III.D and (7)), which entails the following steps.

1) Evaluate the whole dataset $X$ and keep the LRs $Z$ in memory. This has a time complexity of $O(N)$, where $N$ is the number of samples.
2) Build a global $k$-NN index on $Z$ to construct the $k$-NN graph. Using the KD-tree or ball-tree method this has a time complexity of $O(N)$.
3) Run the Louvain algorithm on that graph. Each iteration of the algorithm has a complexity of $O(kN)$, but the number of iterations cannot be known in advance. Since every iteration produces a smaller graph, in the worst case, $N$ iterations are needed for the algorithm to converge, which places the worst-case complexity at $O(kN^2)$.
4) Solve the max-flow-max-weight problem of Section III.C to match the clusters to the true labels. In pactice, since the number of true classes and clusters are negligible compared to $N$, the cost of this step can be ignored.
5) Then, for each true label $t \in L_{\text{true}}$, a new $k$-NN index $\mathfrak{I}_t$ has to be built on the set of CC samples in $t$. If there are $N_{t,\text{ CC}}$ CC samples and $N_{t,\text{ MC}}$ MC samples in $t$, then building $\mathfrak{I}_t$ has an average complexity of $O(N_{t,\text{ CC}})$. The index then needs to be queried $N_{t,\text{ MC}}$ times, which has a complexity of $O(N_{t,\text{ MC}})$. So across all labels, this step has an overall complexity of $O(N_{t,\text{ CC}} + N_{t,\text{ MC}})$. If dataset is roughly balanced (as in the case for classical benchmarks such as CIFAR-100), complexity is $O(N/n_{\text{true}})$, where $n_{\text{true}}$ is the number of true classes.

It is important to note that the cost of computing the Euclidean distance between high-dimensional LRs has to be factored in the cost of steps 2 and 5, which makes even a linear complexity operation very expensive in practice. Furthermore, in step 1., the whole dataset of latent prepresentations $Z$ has to be in memory, which might be prohibitive when the number of samples or the latent dimension is large.

### B. Choice of LS

The efficacy of LCC is expected to be highly dependent on the choice of LS. In general, early layers are designed to extract features form the input samples, and the associated LSs are not expected to contain meaningful clusters. On the other hand, deeper layers are dedicated to classification rather than feature extraction, and it is expected that class-segregated clusters are forming. A preliminary empirical analysis as in Fig. 1 and Fig. 2 can also assist in the choice of LS.

### C. Orthogonal correction

As noted in [5], LRs not only form clusters, but organize along *object manifolds* of lower effective dimension. A MC representation $z$ thus sits in an incorrect manifold $M \subseteq \mathbb{R}^d$. To remedy this, the clustering loss pulls $z$ along a vector $v = z - \bar{z}$, see Section III.D. It could be advantageous to first project $v$ onto $T_z M^\perp$, the subspace orthogonal to the tangent space of $M$ at $z$, in order to accelerate the separation of $z$ from $M$. If $z$ sits in a dense enough region of $M$, approximating the tangent space $T_z M$ can be achieved using principal component analysis, considering the $k'$ closest neighbors of $z$ for a fairly large $k'$.

### D. ImageNet

The ImageNet dataset [29] has become the *de-facto* benchmark for image classification. It is significantly more varied than CIFAR-100 which we considered in this study, but it is also much larger. Considering the computational burden outlined in Section V.A, fine-tuning on ImageNet is not yet practical. Various algorithmic optimizations are under consideration as of the time of writing, see Section VI.

## VI. Possible tradeoffs

### A. Clustering using k-NN pressure

As mentioned in Section V.A, running the Louvain algorithm on the $k$-NN graph of $Z$ has a worst-case complexity of $O(kN^2)$, which in practice makes it the most expensive step in computing the clustering loss. We present *k-NN pressure*, a cheaper method to cluster the dataset of LRs $Z$ that builds on the well-known peer pressure graph clustering method. First, as before, the $k$-NN graph $G = (V, E)$ of $Z$ is constructed. Then, $G$ is clustered as follows.

1) First, each node is placed in its own community, just like with the Louvain method.
2) Then, all nodes are simultaneously moved to the community that is most represented among their neighbors, with ties broken randomly.

3) The node moving step is repeated until the *conductance* of the clustering stops decreasing.

The conductance of a clustering is defined as follows. First, if $C \subseteq V$ is a nonempty subset of nodes, then its conductance is

$$\varphi(C) = \frac{\text{cut}(C, V \setminus C)}{\min(\text{vol}(C, V), \text{vol}(V \setminus C, V))}, \quad (10)$$

where $\text{cut}(D, D')$ is the total weight of edges between $D$ and $D'$, and $\text{vol}(D)$ is the total (weighted) degree of the nodes in $D$. If the denominator is 0, then $\varphi(C) = 1$ by convention. Then, the conductance of a clustering $C_1, ..., C_n$ of $G$ is the weighted mean

$$\Phi = \frac{1}{|V|} \sum_{i=1}^{n} \text{vol}(C_i) \varphi(C_i). \quad (11)$$

The conductance $\Phi$ goes from 0 (perfect clustering) to 1 (worst case).

Since $k \ll N$, it is expected that $G$ is very sparse. The decisive advantage of $k$-NN pressure is to take advantage of this sparsity and operates using efficient sparse matrix operations, see [30] and [31], which significantly alleviates the effective time and memory requirements.

More formally, let $A$ be the (sparse) adjacency matrix of $G$. The algorithm produces a sequence of sparse matrices $B^{(1)}, B^{(2)}, ... \in \mathbb{R}^{N \times N}$, all such that $B^{(r)}_{i,c} = 1$ if at iteration $r$, node $i$ is in cluster $c$.[5] The procedure then goes as follows.
1) We set $B^{(1)}$ to be the identity matrix, meaning that every node starts out in its own cluster.
2) At iteration $r > 1$,
    a) we construct a vector $b \in \mathbb{R}^N$ such that $b_i = \operatorname{argmax}_c AB^{(r-1)}_{i,c}$;
    b) the next sparse matrix $B^{(r)}$ is defined by setting $B^{(r)}_{i,b_i} = 1$;
    c) the volume, cut, and conductance vectors $v, u, \varphi \in \mathbb{R}^N$ of (10) are given by

$$v_j = 2 \sum_i AB^{(r)}_{i,u}$$
$$u = \frac{v}{2} - \text{diag}((B^{(r)})^T AB^{(r)}) \quad (12)$$
$$\varphi = \frac{u}{\min(v, 2N - v + u)},$$

where min is the component-wise minimum;
    d) finally, the conductance $\Phi^{(r)}$ of the clustering is the mean of $\varphi$ weighted by $v$, ignoring the entries where $\varphi$ is undefined (due to some entries in $\min(v, 2N - v + u)$ being 0).

*B. The approximate Louvain loss*

The final step in computing $\mathcal{L}_{\text{clst}}$ is to find a correction target $\bar{z}_i = (z_{j_1} + \cdots + z_{j_k})/k$ for a correctible LR $z_i$. But wether or not $z_i$ is correctible is not known until we try to find enough $z_{j_p}$'s. This requires to:
1) build the $k$-NN index $\mathfrak{I}_{t,C}$ of all CC samples in cluster $C$ matched to true label $t$, provided $C$ has at least $k$ CC samples;

---
[5]Note that there cannot be more than $N$ clusters.

2) for each MC sample $z_i$ in a class $t$, and for each cluster $C$ that has an index $\mathfrak{I}_{t,C}$, query it to find $z^C_{j_1}, ..., z^C_{j_k}$ and compute their centroid $\bar{z}^C_i$;
3) finally, set $\bar{z}_i = \operatorname{argmin}_{\bar{z}^C_i} \|z_i - \bar{z}^C_i\|$.

To partially alleviate this process, we propose an approximate loss that is faster to compute, where instead of considering $k$ "nearest desirable neighbors" $z_{j_1}, ..., z_{j_k}$, we only consider one, possibly not nearest such neighbor.

First, for each cluster $C$, choose a random CC sample $z_C$ in $C$, if such a sample exists. We now say that a sample $z_i$ is *correctible* if there exist at least one cluster $C$ matched to the true label $y_i$ of $z_i$.

The *approximate clustering loss* is defined as

$$\tilde{\mathcal{L}}_{\text{clst}} = \frac{1}{\sqrt{d} N_{\text{corr}}} \sum_{z_i \text{ corr}} \|z_i - \bar{z}_i\|, \quad (13)$$

where

$$\bar{z}_i = \operatorname*{argmin}_{\substack{z_C, \text{ where} \\ C \text{ matched to } y_i}} \|z_i - z_C\|. \quad (14)$$

The process required to compute $\tilde{\mathcal{L}}_{\text{clst}}$ is then:
1) for all cluster $C$, choose a CC sample $z_C$, if possible;
2) for each true label $t$, build the **1-NN** index $\mathfrak{I}_t$ of all the $z_C$'s where $C$ is matched to $t$, if there is at least one such $z_C$;
3) for each MC sample $z_i$ in a class $t$ that has an index $\mathfrak{I}_t$ as above, query $\mathfrak{I}_t$ to find $\bar{z}_i$.

Since the number of clusters matched to a true label $t$ is much smaller than the number of CC samples in $t$, the indices $\mathfrak{I}_t$'s are much smaller, thus faster to build and query. In fact, using exhaustive searches rather than 1-NN indices could be even faster.

## VII. Conclusion

This paper presents a fine-tuning approach that operates on latent spaces by improving the quality and accuracy of latent clusters. The only assumption underpinning our approach is the existence of such clusters, i.e. that during training, a neural network tends to represent similar samples in close proximity, and that high density regions mostly contain representations from the same class. This phenomenon has been observed in every study about the structure of latent spaces. Unlike some of these studies however, we do not make any assumption on the geometry, overall separability, or even the number of such clusters.

We propose a new procedure which we call *latent cluster correction* (LCC) that aims to improve the quality of these clusters. The procedure consists of two steps: first, we use the Louvain community detection algorithm to find these clusters, and then apply a "nearest desirable neighbor" correction loss, which we call the *clustering loss*.

Our preliminary results show that, on average, LCC can noticeable classification accuracy improvements, but at a significant computational cost. It also appeared that the efficiency of LCC is dependent on the architecture of the model under consideration. We expect ongoing optimization efforts to reduce this cost and bring models with larger latent dimen-sions (such as vision transformers) and larger datasets (such as ImageNet) within reach. Furthermore, we plan to apply LCC on deeper versions of ResNet, and investigate the fundamental reason of the lukewarm benchmark results of AlexNet. This latter task is expected to provide further insights into the latent space structures of deep neural networks.